\documentclass[10pt,twocolumn,letterpaper]{article}

\usepackage{cvpr}              

\usepackage{graphicx}
\usepackage{amsmath}
\usepackage{amssymb}
\usepackage[normalem]{ulem}
\usepackage{booktabs}
\usepackage{caption}

\usepackage[breaklinks,colorlinks]{hyperref}

\usepackage[capitalize]{cleveref}
\crefname{section}{Sec.}{Secs.}
\Crefname{section}{Section}{Sections}
\Crefname{table}{Table}{Tables}
\crefname{table}{Tab.}{Tabs.}
\usepackage[tableposition=top]{caption}

\def\cvprPaperID{*****} 
\def\confName{CVPR}
\def\confYear{2023}

\begin{document}

\title{Identity-preserving Editing of Multiple Facial Attributes by Learning Global Edit Directions and Local Adjustments}

\author{Najmeh Mohammadbagheri\\
Amirkabir University\\
Iran\\
{\tt\small n.m.bagheri77@aut.ac.ir}
\and
Fardin Ayar\\
Amirkabir University\\
Iran\\
{\tt\small fardin.ayar@aut.ac.ir}
\and
Ahmad Nickabadi\\
Amirkabir University\\
Iran\\
{\tt\small nickabadi@aut.ac.ir}
\and
Reza Safabakhsh\\
Amirkabir University\\
Iran\\
{\tt\small safa@aut.ac.ir}
}
\maketitle

\begin{abstract}
Semantic facial attribute editing using pre-trained Generative Adversarial Networks (GANs) has attracted a great deal of attention and effort from researchers in recent years. Due to the high quality of face images generated by StyleGANs, much work has focused on the StyleGANs' latent space and the proposed methods for facial image editing. Although these methods have achieved satisfying results for manipulating user-intended attributes, they have not fulfilled the goal of preserving the identity, which is an important challenge. We present ID-Style, a new architecture capable of addressing the problem of identity loss during attribute manipulation. The key components of ID-Style include Learnable Global Direction (LGD), which finds a shared and semi-sparse direction for each attribute, and an Instance-Aware Intensity Predictor (IAIP) network, which finetunes the global direction according to the input instance. Furthermore, we introduce two losses during training to enforce the LGD to find semi-sparse semantic directions, which along with  the IAIP, preserve the identity of the input instance. Despite reducing the size of the network by roughly 95\% as compared to similar state-of-the-art works, it outperforms baselines by 10\% and 7\% in Identity preserving metric (FRS) and average accuracy of manipulation (mACC), respectively. 
\end{abstract}

\begin{figure}
  \centering
   \includegraphics[width=\linewidth]{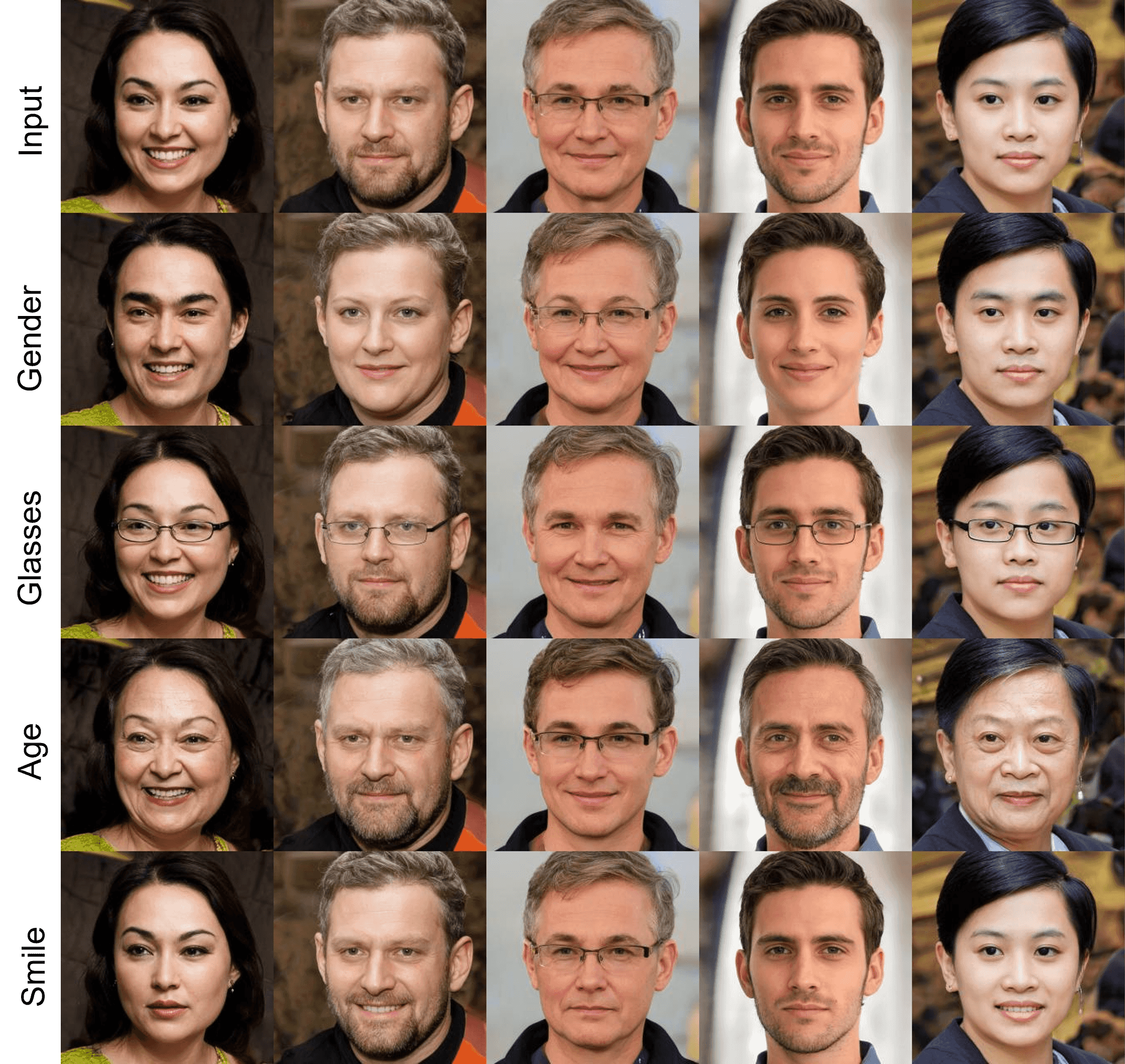}
   \caption{Our model focuses on preserving the identity of the source image during manipulation. Previous works based on pre-trained StyleGAN2 do not maintain the identity features of the source image thoroughly. In this figure, we show the performance of our model. The first row, displays input images to the model. In the following rows, one single attribute is toggled, i.e., if the source image is male, its corresponding manipulated gender attribute is a female face, and vice versa.}
   \label{fig:ours}
\end{figure}

\section{Introduction}
\label{sec:intro}
Generating and manipulating high-resolution and realistic images based on the user intent are among the challenging tasks in computer vision. In the last decade, generative models, including Variational Autoencoders (VAEs) \cite{razavi2019generating}, Generative Adversarial Networks (GANs) \cite{goodfellow2014generative}, Normalizing Flows (NFs) \cite{abdal2021styleflow}, and more recently, Diffusion Models (DMs) \cite{preechakul2022diffusion} have taken valuable steps towards synthesizing realistic images including human face images. However, VAEs and NFs do not generally generate high-quality face images \cite{valenzuela2021expression, li2018facial}, and the sampling process of DMs is far slower than the image generation step of GANs. Consequently, GANs have attracted more attention in face generation and manipulation tasks during the past decade. 

StyleGAN’s series \cite{karras2019style, karras2020analyzing, karras2021alias} are the most celebrated GANs in the realm of facial image generation and have revolutionized the techniques for facial attribute editing as well. These models introduce a learned intermediate latent space, $W$, in which semantic directions are much more disentangled in comparison to the $Z$ space \cite{karras2019style} of the traditional GANs. This means it is possible to find semantic directions in the $W$ space where by moving along each of them, an intended attribute is changed while the other attributes remain intact.  So far, three versions of StyleGAN have been introduced. In this work, we specifically focus on the second version, StyleGAN2 \cite{karras2020analyzing}, to have a fair comparison with state-of-the-art methods. The proposed model is also straightforwardly applicable to the other versions of StyleGAN.

Many researchers have tried to use the intrinsic properties of the StyleGAN’s $W$ space for facial attribute editing. For example, InterfaceGAN \cite{shen2020interfacegan} and GANSpace \cite{harkonen2020ganspace} have noticed the linear separability characteristic of the $W$ space, and by using different statistical methods such as SVMs and PCA, have extracted one global direction for each attribute. Then, they use these global directions for manipulating all instances in the $W$ space, regardless of their location in this space, which leads to some unwanted changes in other attributes.

Some studies \cite{tov2021designing, abdal2019image2stylegan, richardson2021encoding} have also shown that the $W^+$ space consisting of 18 different 512-dimensional vectors is more expressive than the 512-dimensional $W$ space, especially for editing of real images. However, it is remarkable that treating a $W^+$ matrix as a whole, as what DyStyle \cite{li2023dystyle} and ISF-GAN \cite{liu2021isf} have done, is computationally demanding. 

In this paper, we propose a light-weight model for facial attribute editing, named ID-Style, that produces input-specific manipulation directions in the $W$ space of the StyleGAN for each input image. The proposed model properly preserves other attributes of the input image, specifically the identity of the target person, and hence named the ID-Style. To do so, ID-Style comprises three main parts: 1) a Learnable Global Direction (LGD), 2) an Instance-Aware Intensity Predictor (IAIP) network, and 3) an Edit Direction Adjustment (EDA) module. The first part includes M 512-dimensional global directions for M attributes. These global directions form the overall direction of each attribute and are learned during training under some constraints to preserve identity. The goal of the second part is to specify how much each dimension of the $W^+$ should be altered along the global direction so that the desired editing happens and the other attributes remain unchanged. By leveraging a layer embedding, the last component combines the output of the IAIP and the global attribute vectors to create the edit vector for each input image. To further preserve the identity of the input image, two sparsity and direction losses are introduced to enforce the model to find acceptable global directions with the fewest non-zero elements. 

To tackle the challenge of supporting $W^+$ space without enlarging the network’s size, we use the parameter sharing technique in the design of IAIP part of our model, similar to what is done in transformers \cite{vaswani2017attention} or the MLPMixer's architecture \cite{tolstikhin2021mlp}. 

Figure \ref{fig:ours} illustrates the results of the proposed method on some test images. Extensive experiments show that ID-Style brings considerable improvements over state-of-the-art methods, both qualitatively and quantitatively. \\

The contributions of this paper can be summarized as follows: 
\begin{itemize}
    \item Defining sparsity and direction losses to conduct LGD to learn semi-sparse directions. 
    \item Injecting the concept of parameter-sharing into the IAIP model to keep the size of the network small and still support the manipulation of real photos without identity loss. 
    \item Revealing some undiscovered properties of the $W$ latent space about the correlation between identity-preservation and sparseness of semantic directions.
\end{itemize}

The rest of the paper is organized as follows. In Section \ref{sec:related}, the literature on image editing using GANs is reviewed. The proposed method is introduced in Section \ref{sec:method}. In this section, firstly, the intermediate latent space of StyleGAN, $W$ space, is discussed, and then, the ID-Style’s architecture and proposed losses for training it are explained in detail. The superiority of ID-Style over its competitors is exhibited in Section \ref{sec:exp}, through a wide range of experiments. In addition, some analytical experiments on the proposed method and its results are reported in this section.


\section{Related Works}
\label{sec:related}
The manipulation of digital face images has long been a challenging task in computer graphics \cite{khodadadeh2022latent}. Generative models \cite{razavi2019generating, goodfellow2014generative, abdal2021styleflow, preechakul2022diffusion} took a drastic step forward in this way by providing the ability to generate and manipulate images. Even though more recent generative models such as Stable Diffusions \cite{rombach2022high} generate impressive images based on users’ prompts, GANs \cite{goodfellow2020generative} are still more suitable for image attribute manipulation for some reasons: 1) The quality of images generated by GANs is as high as those generated by DMs, 2) The process of synthesizing an image in DMs is far slower than in GANs, and 3) Face images generated by GANs are more realistic than those generated by DMs. So, in this section, we will review GAN-based image manipulation methods. GANs consist of a pair of generator and discriminator models trained jointly in a mini-max game. The generator maps latent space samples to the image space while the discriminator aims to distinguish the generated images from the real ones \cite{voynov2020unsupervised}. GAN-based approaches for image attribute manipulation can be divided into two main categories: 1) image-to-image translation and 2) latent space manipulation. 

Conditional GANs \cite{mirza2014conditional} were introduced to have the ability to control the image generation process and specify attributes of the generated output \cite{brock2018large, he2019attgan}. Image2Image translation models based on conditional GANs are also used to edit the attributes of an image by considering each attribute as a domain \cite{zhu2017unpaired, choi2018stargan, choi2020stargan}. However, since these models take each attribute as a distinct domain, adding more attributes to these models is unhandy. In addition, the resolution of images generated by these models is usually lower than 256×256 due to the computational cost of training a multi-domain GAN for higher-resolution images from scratch. This issue becomes even worse in these models by supporting more attributes and domains \cite{khodadadeh2022latent}.

Another appropriate way for editing attributes of images is to utilize prominent GANs such as PGGAN \cite{karras2017progressive}, StyleGAN \cite{karras2019style}, or StyleGAN2 \cite{karras2020analyzing}, which generate high-resolution images, and find a way to manipulate their outputs. To do so, one has to find directions in the latent space of a pre-trained GAN that the output images exhibit the intended attributes by moving along them. This approach is called latent space exploration (manipulation) \cite{endo2022user}. Supervised and unsupervised methods have been proposed for this purpose.

Unsupervised latent space manipulation methods investigate the latent space using statistical and analytical tools such as principal component analysis (PCA), mutual information, and matrix factorization \cite{jahanian2019steerability, voynov2020unsupervised, yuksel2021latentclr}. GANSpace \cite{harkonen2020ganspace} applies PCA on latent codes to determine interpretable directions among the ones that exhibit the most significant variances. This approach requires a manual selection of desirable semantics. SeFa \cite{shen2021closed} proposes a closed-form factorization algorithm for discovering interpretable directions in a latent space by decomposing the weights of a pre-trained GAN. In LowRankGAN \cite{zhu2021low}, a low-rank factorization algorithm is applied to analyze semantics in a latent space and find steerable directions. Even though these approaches do not need any out-of-shelf model to explore latent space and generate realistic manipulated images, they often entangle multiple attributes together and suffer from unwanted changes of attributes in their outputs. 

On the other hand, there are supervised methods for latent space manipulation that study the latent space of pre-trained GANs using auxiliary classifiers of attributes. InterFaceGAN \cite{shen2020interfacegan} applies a support vector machine (SVM) to classify each facial attribute in the latent space of a pre-trained GAN and accomplishes attribute manipulation by moving along the normal direction of the corresponding separating hyperplane. While this method reveals some characteristics of the latent space of StyleGAN, its two shortcomings are the vast number of positive and negative data needed for each attribute and unwanted changes in global attributes when manipulating local features. StyleSpace \cite{wu2021stylespace} performs a quantitative study on the style space of StyleGAN2 and discovers a highly localized and disentangled control of the visual attributes by focusing on style channels. However, InterfaceGAN and StyleSpace find a general direction for each attribute and their semantic directions are independent of the location of the input vector in the latent space of StyleGAN. Notably, it is impossible to edit multiple attributes at once in these methods, and multiple single-attribute edits have to be done for this purpose.

 Some works utilize the information of the input latent vector and its location on the manifold of the meaningful points in this space to find a semantic direction. StyleFlow \cite{abdal2021styleflow} proposes a facial attribute editing model using conditional continuous normalization flows. ISF-GAN \cite{liu2021isf} trains a network with an MLP architecture for latent vector transformation in the $W^+$ space of StyleGAN2, which edits images in a multi-domain and multi-modal manner. DyStyle \cite{li2023dystyle}, GuidedStyle \cite{hou2022guidedstyle}, and the model proposed by Wang et al. \cite{wang2021attribute} are other similar works to ISF-GAN which utilize neural networks with different architectures and formulation to modify $W^+$ instance vectors to represent face images with/without specified attributes. Even though the results of these methods are more precise than previous ones, the sizes of the models are massive and they have high computational costs and need substantial resources to be trained.

 In this paper, we propose a supervised facial attribute editing model that tackles the problem of attribute entanglement in the $W$ space of StyleGAN2 so that there are no unwanted changes in the results. Besides, the proposed model is far smaller than existing models because of its compact architecture inspired by some intrinsic properties of $W^+$ space. Consequently, re-training and adding more attributes to the proposed model are much more convenient than the previous works. In addition, it can simultaneously edit multiple attributes of a high-resolution face image in real time.

\section{The Proposed Method}
\label{sec:method}
This section introduces the proposed facial attribute editing model (ID-Style) and the motivations behind its architecture. In the following, the intermediate latent space of StyleGAN is briefly introduced in Section \ref{subsec:latentspace}. Then, the overall structure of ID-Style is outlined in Section \ref{subsec:outline}. Section \ref{subsec:arch} details each ID-Style’s constituent. Then, in Section \ref{subsec:loss}, loss functions used in the training phase are described. Finally, Section \ref{subsec:infer} discusses how to manipulate multiple attributes using ID-Style simultaneously.
\subsection{Intermediate Latent Space of StyleGAN}
\label{subsec:latentspace}
All StyleGAN versions sample from a learnable latent space called $W \in R^{512}$ to determine the various styles needed to create an image. Put succinctly, every aspect of an image, ranging from the color of pixels to high-level semantic attributes like age (in the case of face images), is encoded into a $w \in R^{512}$ vector of $W$. For both StyleGAN1 \cite{karras2019style} and StyleGAN2, the sampled latent vector $w$ traverses all 18 layers of the generator network during the image synthesis process.

According to the original StyleGAN paper \cite{karras2019style}, the input latent vector for each of the 18 layers of the generator can differ, as the input of each layer is more pertinent to certain parts of the output image. This brings us to the $W^+ \in R^{18\times 512}$ space in which each instance is a matrix consisting of 18 different $w$ vectors. Also, as the $W^+$ space has a higher degree of freedom, it allows better encoding of real images into the latent space of StyleGANs.

Each point of the latent space of StyleGAN is converted to an image by the generator part of this model. Any changes in this latent point will affect some attributes of the generated image. The target of image editing by manipulating the latent space of StyleGAN is to find a direction in $W^+$ (or $W$) that the target attribute would change in a desired way by moving along that direction. In this paper, we focus on facial attribute manipulation in $W^+$ space of StyleGAN2. 

\subsection{Outline }
\label{subsec:outline}
The inputs of ID-Style are a $w^+\in W^+$ vector sampled from the latent space of StyleGAN2 or calculated for a real image, and a target attribute vector $attr\in \{-1, 0, 1\}^M$ for $M$ different facial attributes. In the case of real photo editing, an auxiliary module - an encoder or an optimization procedure- is used to map the input image to its corresponding latent code in the $W^+$ space. The attribute vector specifies the target value of each attribute: 1 to add (increase) the attribute (for example, to add a smile or increase the age), -1 to remove (decrease) the attribute, and 0 to keep the attribute unchanged. The proposed model can be used for both single-attribute editing and multi-attribute editing. In the single-attribute mode, ID-Style outputs $M$ edited images according to the values of the $M$ elements of the attribute vector. In the multi-attribute mode, only one image is returned that contains all requested attribute values (described in Section \ref{subsec:infer}).

ID-Style consists of three components: a set of Learnable Global Directions (LGD), an Instance-Aware Intensity predictor (IAIP), and an Edit Direction Adjustment (EDA) module. The first component is responsible for finding $M$ global semantic directions ($P_{m} \in R^{512}$) during training. In the inference time, these directions are fixed for all input instances. The IAIP module is used to fine-tune global directions regarding the locations of the input instances in the intermediate latent space. The input of IAIP is a $w^+$ concatenated with positional embedding (described in the following), and its output is $l_{f t} \in R^{18 \times 512}$ used to adjust the global directions. The last component uses the output of the IAIP, global edit directions, and the input attribute vector to produce the final edit vector for each input image.

\subsection{Architecture}
\label{subsec:arch}
\begin{figure*}[t]
  \centering
    \includegraphics[width=\linewidth]{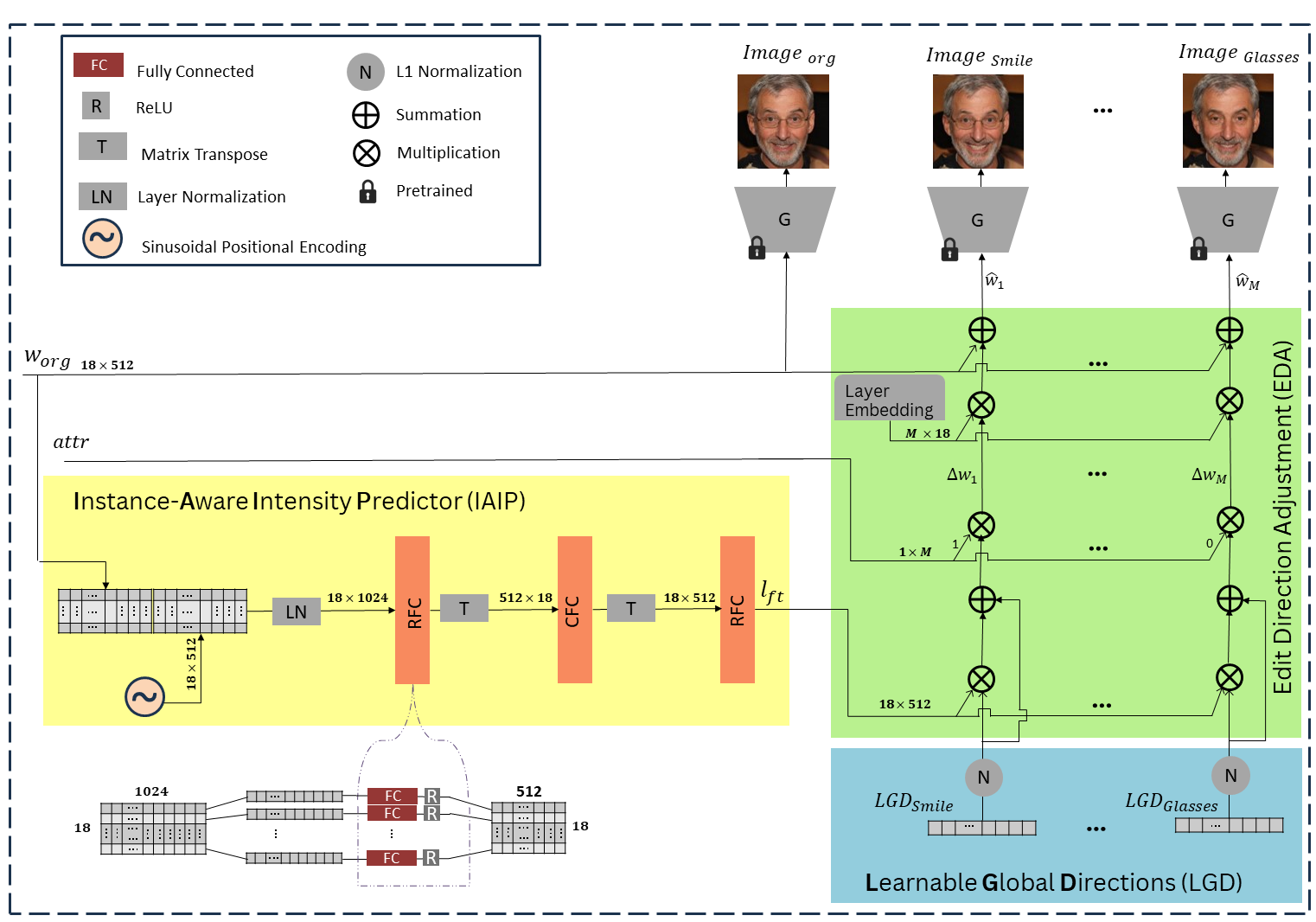}
   \caption{
   The overall architecture of ID-Style. The proposed model learns to predict $M$ semantic edit vectors for each $w_{org} \in W^+$. These edit vectors are added to input $w_{org}$ in order to alter $M$ facial attributes of it. This process is factorized in three steps: 1) learning a global instance-agnostic direction in \textit{Learnable Global Direction}, 2) predicting the fine-tuned coefficients for each instance through \textit{Instance-Aware Intensity Predictor}, and 3) adjusting the global direction based on the output of the IAIP module, the \textit{layer embedding}, and the target attribute vector $attr$. The final edited images are obtained by adding the edit vector to the initial $w_{org}$ and reconstructing the images using the pre-trained generator ($G$).}
   \label{fig:arch}
\end{figure*}

The detailed architecture of the ID-Style is given in Figure \ref{fig:arch}. The proposed network accepts a $W^+$ instance $w_{org}\in R^{18\times 512}$ and the target values of $M$ facial attributes as $attr \in \{-1,1\}^M$, and produces a set of $\Delta w_m \in R^{18\times 512}, m \in \{1,...,M\}$, each of which for altering one attribute of the input $w_{org}$. The edit vector $\Delta w_m$ is added to $w_{org }$ to generate the edited latent vector $\widehat{w}_{m}$. As mentioned earlier, the Learnable Global Direction (LGD) comprises a set of learnable parameters $P_{m} \in R^{512}$ each representing a global direction for editing an attribute, like the global directions of InterfaceGAN \cite{shen2020interfacegan}. The global directions of LGD are instance-agnostic and L1-normalized directions which are subsequently fine-tuned by the output of the second part of ID-Style. LGD's edit vectors are normalized to ensure they focus on learning the true directions of different edits, not the exact amount of movement along those directions. This facilitates the learning of the second component of the model (IAIP), which is responsible for determining the exact intensity of moving along each dimension of the global direction for each $W^+$ instance; hence it is named instance-aware intensity predictor (IAIP).

Input of IAIP is a $w^+$ matrix. As mentioned in Section \ref{sec:intro}, unlike what is done in previous works (e.g. \cite{li2023dystyle, liu2021isf}), we treat the input $w_{org}$ as a sequence of 18 distinct 512-dimensional $w$ vectors. So, similar to sequence modeling architectures, like transformers\cite{vaswani2017attention} and MLPMixer \cite{tolstikhin2021mlp}, we use parameter sharing to separately process each of these 18 vectors. We employ an MLPMixer-like architecture to maintain the size of the network small, in addition to utilizing all information of the input matrix. Although the $w$ vectors (rows of $w^+$) are the same in nature, they go to different layers of StyleGAN2, and therefore produce a different behavior in the output of the StyleGAN2. Thus, it is crucial to add layer-specific information to each $w$ vector, similar to what is done in transformers. Our approach involves concatenating a sinusoidal positional encoding ($PE$) \cite{vaswani2017attention} to each row of $w_{org}$.

The overall $PE$ has the same dimension as that of $w_{org}$ and enables IAIP network to incorporate layer-specific information and effectively differentiate between individual $w$ vectors within the $w_{org}$.  After this, the resulting matrix goes through the IAIP network, whose structure has been inspired by the MLPMIixer’s architecture \cite{tolstikhin2021mlp}. Although IAIP module is position-variant, we found that this extra level of position information is still helpful. The IAIP comprises two Row-wise Fully-Connected(RFC) layers that extract layer-specific features of the 18-layered input instances and one Column-wise FC(CFC) layer that enables the exchange of information between these layers:

\begin{equation}
\begin{gathered}
l_{f t}=R F C_{512 \times 512} ( C F C_{18 \times 18}( \\ R F C_{1024 \times 512} (w_{org} \| P E)))
\end{gathered}
\end{equation}
with
\begin{equation}
\begin{gathered}
R F C_{a \times b}(x)=\operatorname{ReLU}\left(F C_{a \times b}(x)\right) \\
C F C_{a \times b}(x)=\operatorname{ReLU}\left(F C_{a \times b}\left(x^T\right)^T\right)
\end{gathered}
\end{equation}
where $\|$ is the concatenation operation, $FC_{a\times b}$ is an FC layer with input and output dimensions of a and b, respectively, and $T$ stands for transposing the last two dimensions of an input tensor. $l_{f t}\in R^{18\times 512}$ is the output of IAIP which fine-tunes the global directions of the LGD.
\\
This two-part structure is inspired by the latent space properties of StyleGANs, as we elucidated in Section \ref{sec:intro}. The final output of our network for the $m$th attribute ($\widehat{w}_{m}$) is computed in the Edit Direction Adjustment module as follows. At first, the edit vectors are calculated as 
\begin{equation}
\Delta w_{m}^i=l_{f t}^i \odot P_{m}+P_{m}, \quad i \in\{1,...,18\}, m \in\{1,...,M\}
\end{equation}
where $\Delta w_{m}^i$ is the $i$th row of the edit matrix generated for the $m$th attribute, and $\odot$ is element-wise multiplication. Note that the output of IAIP is invariant to the number of attributes as we guide (based on our experiments) the learnable parameters $P_{m}$ in LGD to be sparse vectors. Therefore, each element of the IAIP output corresponds only to a single attribute. In the next section, we discuss how we guide the network to learn directions in this manner.

Up to this point, our network is unaware of target face attributes $attr$. Assuming the relative disentanglement of attributes in StyleGAN’s latent space, we can easily inject our desired target facial attributes into the network by:
\begin{equation}
\Delta w_{m}={attr}_{m} \cdot \Delta w_{m}
\end{equation}
This equation simply assumes that the direction of adding an attribute is the opposite of the direction of removing that attribute.

As noted by StyleFlow \cite{abdal2021styleflow} and adopted by ISF-GAN \cite{liu2021isf}, it is inadvisable to modify all layers of StyleGAN2's latent space (i.e., each row of $w_{org}$) in the same manner, as editing the latter layers may lead to unnecessary changes of the target image, potentially causing the loss of identity. To circumvent this issue, we introduce an additional embedding layer to learn how much of the $\Delta w$ should be incorporated into each layer of  the input instance $w_{org}$:
\begin{equation}
\widehat{w}_{m}^i=w_{o r g}^i+\operatorname{sigmoid}\left(E_{m}^i\right) \cdot \Delta w_{m}^i, \quad i \in\{1,...,18\} 
\end{equation}

In the above equation, $E_{m} \in R^{18}$ is a learnable embedding for attribute $m$, that determines the amount of changes of each layer of $\Delta w_{m}$. $\widehat{w}_{m}$ is the latent matrix of the edited face image in which the $m$-th attribute is altered.

\subsection{Training}
\label{subsec:loss}
The proposed ID-Style model is trained through a set of loss functions described in this section. As mentioned earlier, we want the learnable parameters $P_{m}$s in LGD to be sparse. To achieve this goal, a sparsity loss is used in the training of the model, which is defined as: 
\begin{equation}
\mathcal{L}_{sparsity }=\sum_{m}\left\|P_{m}\right\|_1
\end{equation}
Since the output of IAIP network is unbounded and could be any positive value, the learnable parameters ($P_{m}$s) may not necessarily learn a global direction for each attribute. This is because the IAIP’s output has the capacity to arbitrarily alter the direction of the learned global directions. Therefore, in order to mitigate this issue, we employ an additional loss function referred to as direction loss. Specifically, we define the direction loss as the cosine distance between the fine-tuned direction and the global direction for each input as follows:
\begin{equation}
\mathcal{L}_{direction }=\sum_{m=1}^{M} \sum_{i=1}^{18}\left(1-\operatorname{sim}\left(l_{f t}^i \odot P_{m}, P_{m}\right)\right)
\end{equation}
where $\operatorname{sim}$ stands for cosine similarity. The previous research has shown \cite{liu2021isf}, and we also found that using an extra neighborhood loss is helpful in order to keep the edited and original input close to each other:
\begin{equation}
\mathcal{L}_{nb}=\sum_{m}\left\|\widehat{w}_{m}-w_{org}\right\|_2
\end{equation}

Finally, each $\widehat{w}_{m}$ is given to a StyleGAN2 synthesis network to generate the corresponding face image $I_{m}$. For the generated images to have the target attributes ($attr$), the model is trained by a multi-target binary cross entropy loss $\mathcal{L}_{\text {class }}$ using pre-trained attribute classifiers \cite{liu2021isf}: 

\begin{equation}
\mathcal{L}_{\mathrm{class}}=-\log D(\boldsymbol{attr} \mid G(E(\boldsymbol{w^+}, \boldsymbol{attr})))
\end{equation}
where $D(.)$ is a pre-trained attribute classifier, and $E(.)$ represents the output of ID-Style. 

To preserve original content and face identity, we use a cosine similarity loss $\mathcal{L}_{id}$ in the feature space of an identity-aware network, ArcFace \cite{deng2019arcface} as a feature extractor model.
\begin{equation}
\mathcal{L}_{\mathrm{id}}=1-\text{sim}\left(\psi(G(w^+)), \psi(G(E(\boldsymbol{w^+}, \boldsymbol{attr})))\right) .
\end{equation}
where $\psi(.)$ refers to feature maps extracted by the ArcFace \cite{deng2019arcface}. 

The full objective function can be summarized as follow:
\begin{equation}
\begin{gathered}
\mathcal{L}_{\text{total}} = \lambda_{\text{class}} \mathcal{L}_{\text{class}}+\lambda_{\text{nb}} \mathcal{L}_{\text{nb}}+\lambda_{\text{sparsity}} \mathcal{L}_{\text{sparsity}} \\
+\lambda_{\text {direction}} \mathcal{L}_{\text {direction}}+\lambda_{\text{id}} \mathcal{L}_{\text{id}}
\end{gathered}
\end{equation}
where $\lambda_{\mathrm{class}}, \lambda_{\mathrm{nb}}, \lambda_{\mathrm{sparsity}}, \lambda_{\mathrm{direction}},$ and $\lambda_{\mathrm{id}}$ are the hyperparameters for each loss term.
\\
\subsection{Multi Attribute Inference} 
\label{subsec:infer}
Similar to the training step, the inference is made by providing the network with a $w^+$ instance along with the target labels $attr \in \{-1,0,1\}^M$. ID-Style provides $\Delta w_m (m=1,2,...,M)$ for altering each of $M$ attributes. During training, we do not train our network with multi-attribute editing, but as our network is guided to preserve the identity and be sparse in terms of semantic directions, we can use it in multi-attribute editing by a simple heuristic:
\begin{equation}
\Delta w^{i, j}=absmax _{m}\left(\Delta w_{m}^{i, j}\right)
\end{equation}
where $absmax$ returns the value whose absolute is the largest value among its input set.
The above equation states that for simultaneously editing a list of attributes of $w_{org}$, we simply use the maximum value along all $\Delta w_{m}$s.

\section{Experiments}
\label{sec:exp}
\begin{figure*}[t]
  \centering
  \includegraphics[width=\linewidth]{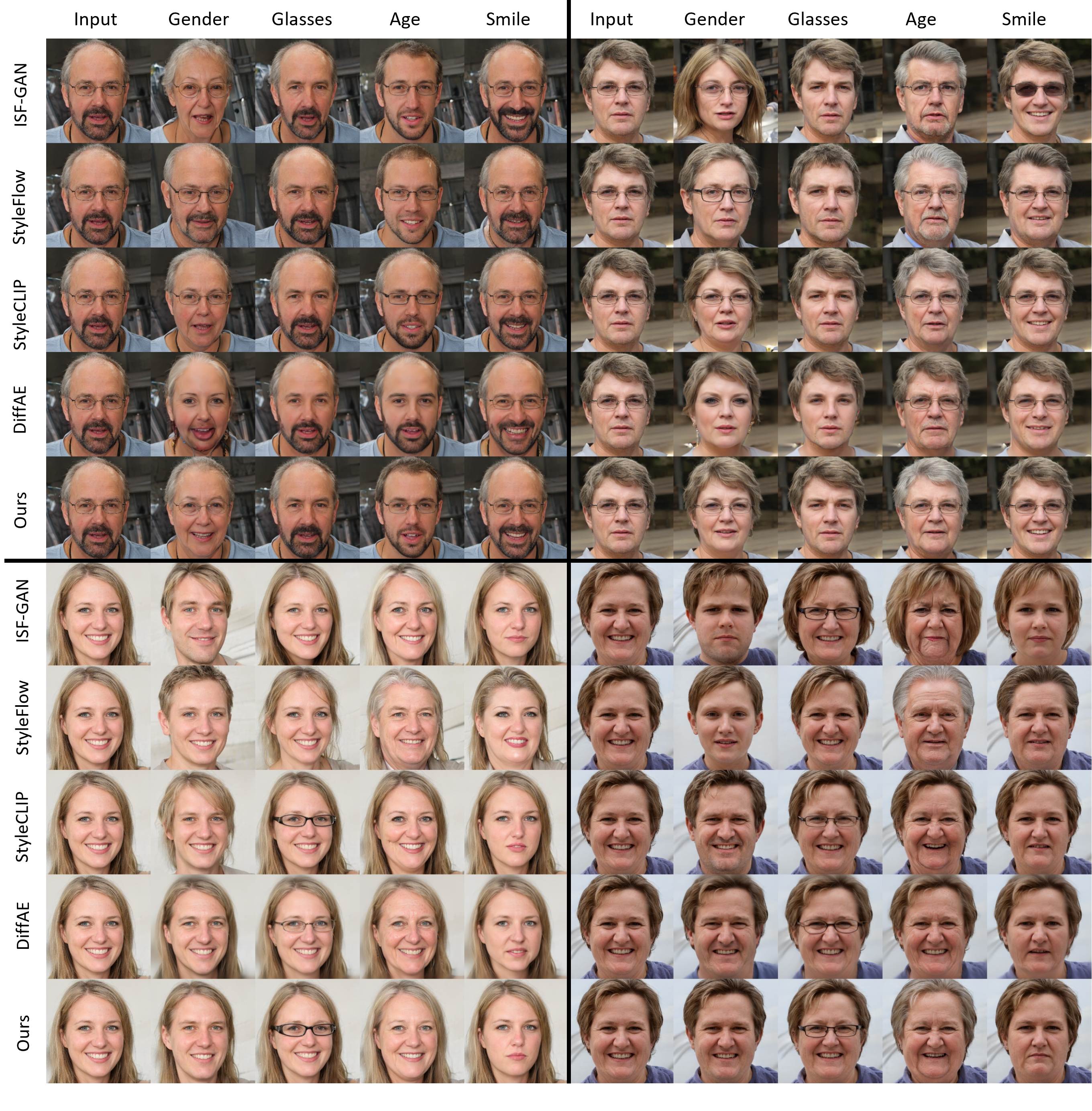}
   \caption{Visual Comparison between state-of-the-art methods and ID-Style on four different input images. ISF-GAN \cite{liu2021isf}, StyleFlow \cite{abdal2021styleflow}, and StyleCLIP \cite{patashnik2021styleclip} are based on pre-trained StyleGAN2, similar to ours. 
   To have a fair judgment, the experiment performed on the $W^+$ space of stylegan2. DiffAE \cite{preechakul2022diffusion} has a completely different architecture and is based on diffusion models.} 
   \label{fig:complete}
\end{figure*}

\begin{table*}
  \centering
  \small
  \caption{Quantitative comparisons on identity and content preservation, accuracy of editing, and image quality of manipulated images.}
  \label{tab:main_quant}
  \begin{tabular}{c c c c c c | c c c c}
    \toprule
    Method & $\uparrow SSIM$ & $\uparrow FRS$ & $\downarrow FID$ & $\downarrow LPIPS$ & $\uparrow mACC(\%)$ & $Gender(\%)$ & $Glasses(\%)$ & $Age(\%)$ & $Smile(\%)$ \\
    \midrule
    ISF-GAN(\textit{W}) & 0.626 & 0.653 & \textbf{20.67} & 0.153 & 88.55 & 89.4 & 94.9 & 76.2 & 93.6 \\
    InterfaceGAN(\textit{W}) & 0.607 & 0.664 & 23.57 & 0.156 & 86.45 & 82.1 & 86.7 & 77.5 & 99.3 \\
    Styleflow(\textit{W}) & 0.650 & 0.640 & 45.13 & 0.20 & 72.41 & 70.2 & 61.7 & 82.6 & 75.0 \\
    Ours(\textit{W}) & \textbf{0.686} & \textbf{0.757} & 26.33 & \textbf{0.105} & \textbf{96.73} & \textbf{94.9} & \textbf{97.7} & \textbf{94.6} & \textbf{99.6}\\
    \midrule
    ISF-GAN($\textit{W}^+$) & 0.622 & 0.644 & \textbf{22.52} & 0.157 & 89.25 & 90.9 & 96.2 & 76.6 & 93.2 \\
    Styleflow($\textit{W}^+$) & 0.641 & 0.622 & 48.43 & 0.209 & 73.72 & 72.9 & 62.2 & 84.7 & 75.0 \\
    Ours($\textit{W}^+$) & \textbf{0.685} & \textbf{0.761} & 28.80 & \textbf{0.105} & \textbf{96.67} & \textbf{94.9} & \textbf{98.2} & \textbf{94.1} & \textbf{99.4}\\
    \bottomrule
  \end{tabular}
\end{table*}

In this section, we compare the performance of the proposed model with state-of-the-art models based on StyleGAN2 \cite{karras2020analyzing} pre-trained on the FFHQ dataset \cite{karras2019style} and a Diffusion-based model \cite{preechakul2022diffusion}. All experiments conducted on StyleGAN2 to have a fair comparison with other GAN-based methods. Some analytical experiments are also designed to investigate the correctness of the assumptions underlying the proposed model. 

\textbf{Dataset.} The experiments of this section use the dataset released by StyleFlow \cite{abdal2021styleflow} which is sampled from $W$ space and consists of 10K training and 1K testing latent vectors for StyleGAN2 with annotated attributes. For each latent code, similar to ISF-GAN \cite{liu2021isf}, we randomly select another latent code and then permute these two latent codes to generate a new one in $W^+$ space. This procedure is repeated 8 times for each sample in the training set and 1 time for the test set, resulting in a final training dataset of 80K latent vectors. 

\textbf{Training Details.} Our model is trained for 100K iterations with a batch size of 2 on a single 1080Ti GPU for 24 hours with our implementation in PyTorch \cite{paszke2017automatic}. Compared to the state-of-the-art models, our model has far fewer learnable parameters ( see Table \ref{tab:param}), enabling the model to be trained on a single GPU with 12GB RAM. The weight parameters $\mathcal{L}_{nb}$, $\mathcal{L}_{direction}$, $\mathcal{L}_{sparsity}$, $\mathcal{L}_{class}$, and $\mathcal{L}_{id}$ are set to 0.3, 1, 1, 2, and 5, respectively, based on our experiments. The AdaBelief \cite{zhuang2020adabelief} optimizer with $\beta_1=0.98$ and $\beta_2=0.98$ is used. The learning rate is set to 0.001.

\begin{table}
  \centering
  \small
  \caption{The number of learnable parameters of different models and the average time each model takes to manipulate a latent code.}
  \label{tab:param}
  \begin{tabular}{c c c }
    \toprule
    Method & \# Parameters & Inference Time (ms) \\
    \midrule
    ISF-GAN & 16,327,168 & $\sim$ 26  \\
    Styleflow & 1,691,649 & $\sim$ 621 \\
    Ours & \textbf{793,946} & \textbf{$\sim$ 2 } \\
    \bottomrule
  \end{tabular}
\end{table}

\textbf{Evaluation Metrics.} Two main targets of the face manipulation models should be the ability to create the desirable attributes in the output image (accuracy) and preserve the identity of the input image. For the former, we calculate the average accuracy of manipulation over 1k images using off-the-shelf classifiers \cite{karras2019style}. For each image, four different attributes are manipulated for both -1 (removal) and 1 (addition) directions, resulting in a total of 8k manipulated images. To measure the identity and content preservation performance, we utilize a wide range of metrics to have a comprehensive comparison. SSIM \cite{wang2004image} measures the local structure similarity between the input and manipulated images. The higher SSIM value, the more content is preserved. FRS \cite{liu2021isf} estimates the cosine similarity between features extracted from two facial images using the ArcFace network \cite{deng2019arcface}. LPIPS \cite{zhang2018unreasonable} represents the perceptual distance between the input and output images. A greater value for FRS and LPIPS means better identity preservation. We also calculate FID \cite{heusel2017gans} for the diversity and quality of the manipulated images. All of the above metrics are computed on the aforementioned test set consisting of 1k input images and 8k output images. 

\textbf{Baselines.} We compare ID-Style with three state-of-the-art methods similar to ours in employing a pre-trained and fixed unconditional GAN to manipulate faces and domains of attributes. These methods are InterFaceGAN \cite{shen2020interfacegan}, StyleFlow \cite{abdal2021styleflow}, and ISF-GAN \cite{liu2021isf}. We also compare the results of the proposed model with those of the state-of-the-art methods StyleClip \cite{patashnik2021styleclip} and DiffAE \cite{preechakul2022diffusion}. The former is a text-driven model, and the latter is a Diffusion-based model. The original and officially released codes are used for all baseline methods in the experiments. 

It should be noted that the above baseline modes use different latent spaces for image manipulation.  StyleFlow \cite{abdal2021styleflow} and ISF-GAN \cite{liu2021isf} manipulate images in the $W^+$ space. InterfaceGAN \cite{shen2020interfacegan} performs changes in the $W$ space. The best version of StyleCLIP \cite{patashnik2021styleclip} uses the $S$ space of styleGAN2. Finally, DiffAE \cite{preechakul2022diffusion} has a completely different latent space and approach. The official code of all baselines is based on StyleGAN2 \cite{karras2020analyzing} pre-trained on FFHQ \cite{karras2019style}, except InterfaceGAN \cite{shen2020interfacegan} whose official hyperplanes are from the StyleGAN \cite{karras2019style} latent space. To have a fair comparison, we extract separator hyperplanes for four facial attributes in the latent space of StyleGAN2 using the official code of InterfaceGAN \cite{shen2020interfacegan}. 

Since DiffAE \cite{preechakul2022diffusion} and StyleCLIP \cite{patashnik2021styleclip} only specify the direction of attribute control and the intensity of manipulation should be set manually, we have tried to find the best coefficient of moving for each sample and attribute for these two models in the qualitative comparison. Also, because of this issue, we do not compute numerical evaluation metrics for these models. InterfaceGAN \cite{shen2020interfacegan} also needs the coefficient of moving along the attribute vector, but it is not as sensitive as previous methods. Thus we simply used the values of 3 and -3 for this model coefficients. To provide a more comprehensive evaluation, we compared the proposed method with the other models in both $W$ and $W^+$ spaces while ID-Style has been trained in the $W^+$ space. 

\subsection{Qualitative Comparison}

Figure \ref{fig:complete} portrays the visual performance of different methods through 4 random face images generated by StyleGAN2 \cite{karras2020analyzing}. Four attributes of each sample, namely Gender, Glasses, Age, and Smile, are toggled using five facial attribute editing models. It is evident from the results that ISF-GAN \cite{liu2021isf} changes the identity features during gender manipulation and also alters other attributes, such as the shape of glasses and hair.

Likewise, StyleFlow \cite{abdal2021styleflow} changes the identity of the input face images. Moreover, it does not apply the desirable edits in some cases e.g., the glasses attribute. StyleCLIP \cite{patashnik2021styleclip} and DiffAE \cite{preechakul2022diffusion} are better than the two previous methods in identity preservation. However, our model is more precise than DiffAE and StyleCLIP's in preserving identity. In addition, the main drawback of these two models is that they need a tremendous effort to find the best intensity for each attribute. It is also notable that there are some artifacts in the DiffAE’s outputs. On the contrary, the identity of the input image in all cases is clearly preserved during manipulation using the proposed method. Besides that, the proposed model has applied the intended manipulation much faster than StyleCLIP and DiffAE’s. 

\begin{figure}
  \centering
   \includegraphics[width=\linewidth]{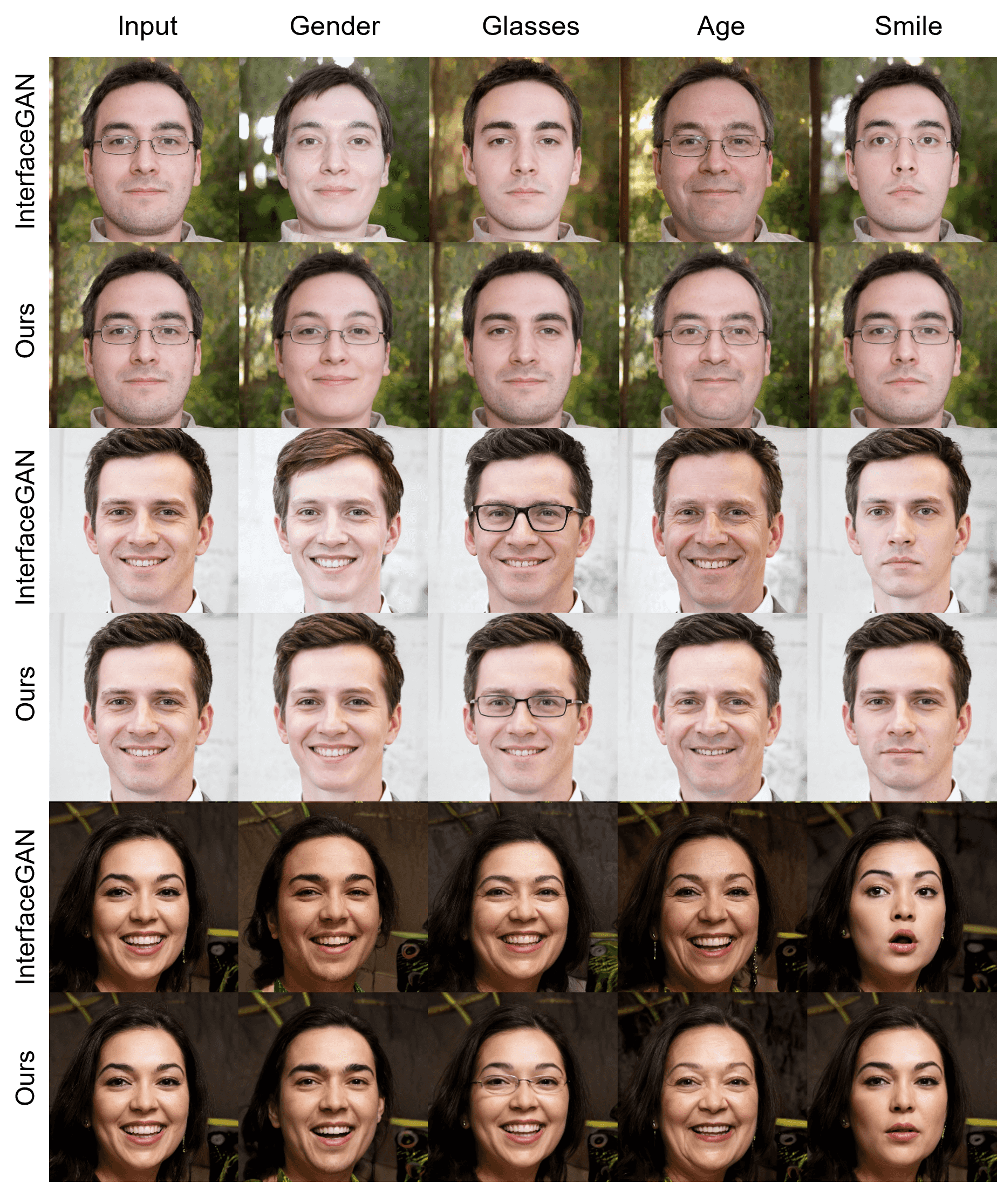}
   \caption{Qualitative Comparison between InterfaceGAN \cite{shen2020interfacegan} and ID-Style in the $W$ space of StyleGAN2.}
   \label{fig:wspace}
\end{figure}

The results of InterfaceGAN \cite{shen2020interfacegan} and the proposed method are compared in the figure \ref{fig:wspace}. In this experiment, both models use the $W$ space for attribute manipulation. It is quite obvious that our method outperforms InterfaceGAN in both content preservation and attribute manipulation.

\begin{figure}
    \centering
    \includegraphics[width=\linewidth]{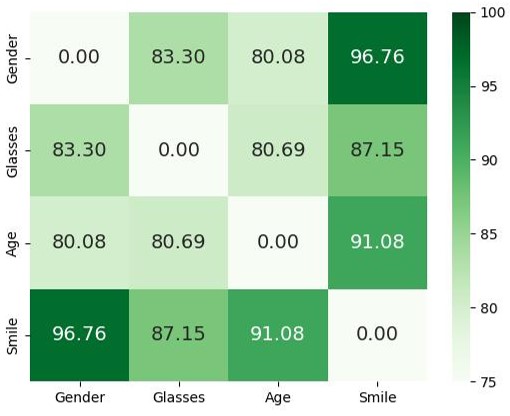}
    \caption{Angle between every two Global Directions in degree.}
    \label{fig:degre}
\end{figure}

\subsection{Quantitative Comparison}
The performance of different methods in both $W$ and $W^+$ spaces in terms of numerical evaluation metrics is reported in Table \ref{tab:main_quant}. The FRS and LPIPS of the manipulated images show that the proposed model properly preserves the identity compared to other methods. The SSIM results indicate that content preservation in ID-Style is better than other methods. In addition, the precision of controlling the target attributes (mACC) of the proposed method is higher than those of the other approaches. While FID of ISF-GAN is lower than the proposed model, the visual comparison of the results of the two models does not show any superiority for ISF-GAN, and the outputs of the proposed model are visually realistic and have a high quality and resolution ($1024 \times 1024$). It is worth mentioning that FID does not reflect human judgments in the quality of generated images \cite{tov2021designing}. The detailed accuracies for the four studied attributes are also given in Table \ref{tab:main_quant}.

\subsection{Multi Attribute Editing}
\begin{figure}
  \centering
   \includegraphics[width=\linewidth]{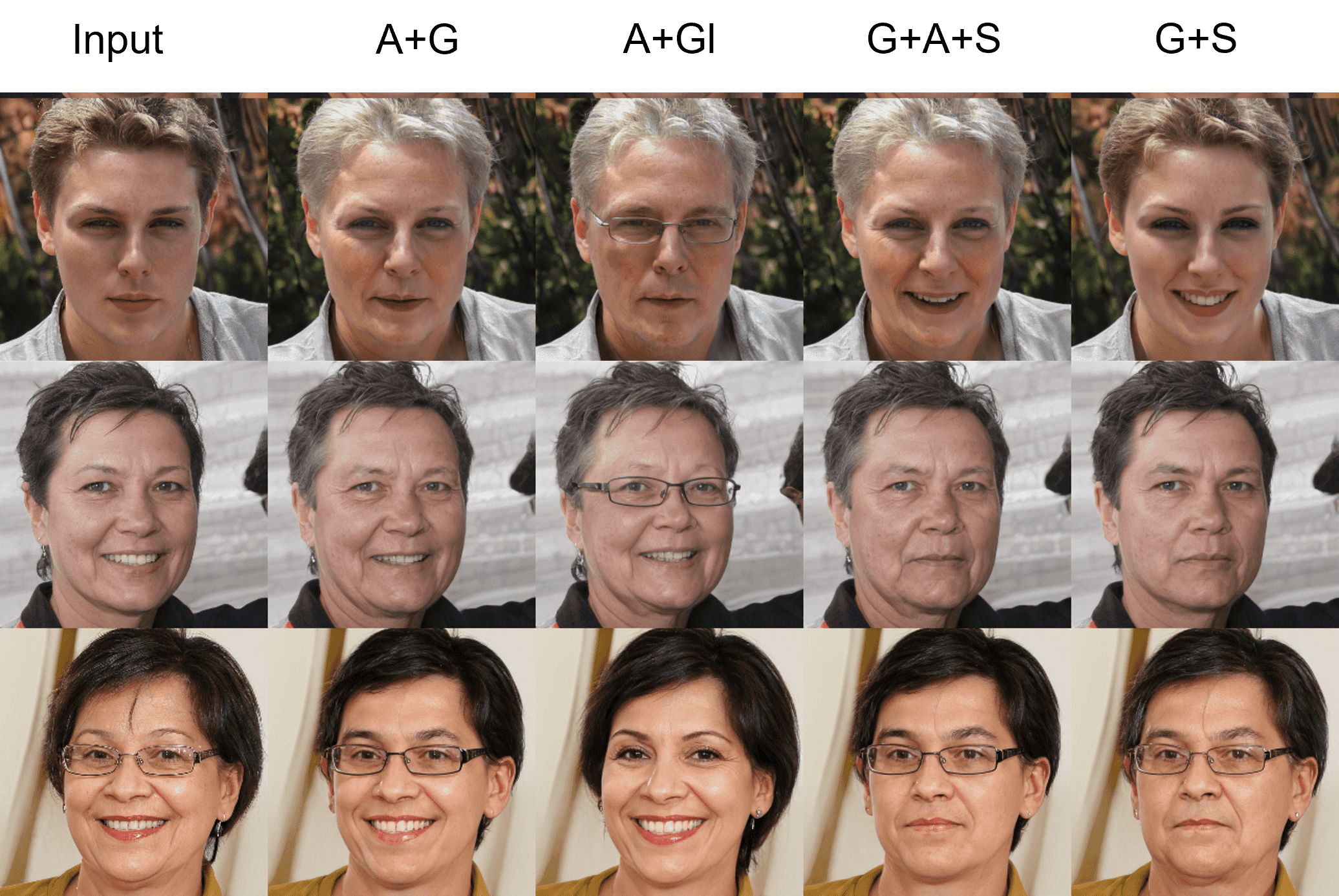}
   \caption{Our model's performance in multi-attribute editing, where “G”, “Gl”, “A” and “S” refer to Gender, Glasses, Age, and Smile attributes, respectively.}
   \label{fig:multi}
\end{figure}

As mentioned in Section \ref{sec:method}, we do not train ID-Style in multi-attribute editing mode, but due to the architecture of ID-Style and the sparsity of attribute directions learned by this model, it is possible to simultaneously manipulate multiple attributes using ID-Style. In other models such as InterfaceGAN \cite{shen2020interfacegan}, multiple attribute editing is realized via repeatedly applying single attribute editing to the output of the previous manipulation step, a process that worsens identity loss in these models. Figure \ref{fig:multi} indicates how well the proposed method carries out multi-attribute editing.

\subsection{Real Image Editing}
\begin{figure}
  \centering
   \includegraphics[width=\linewidth]{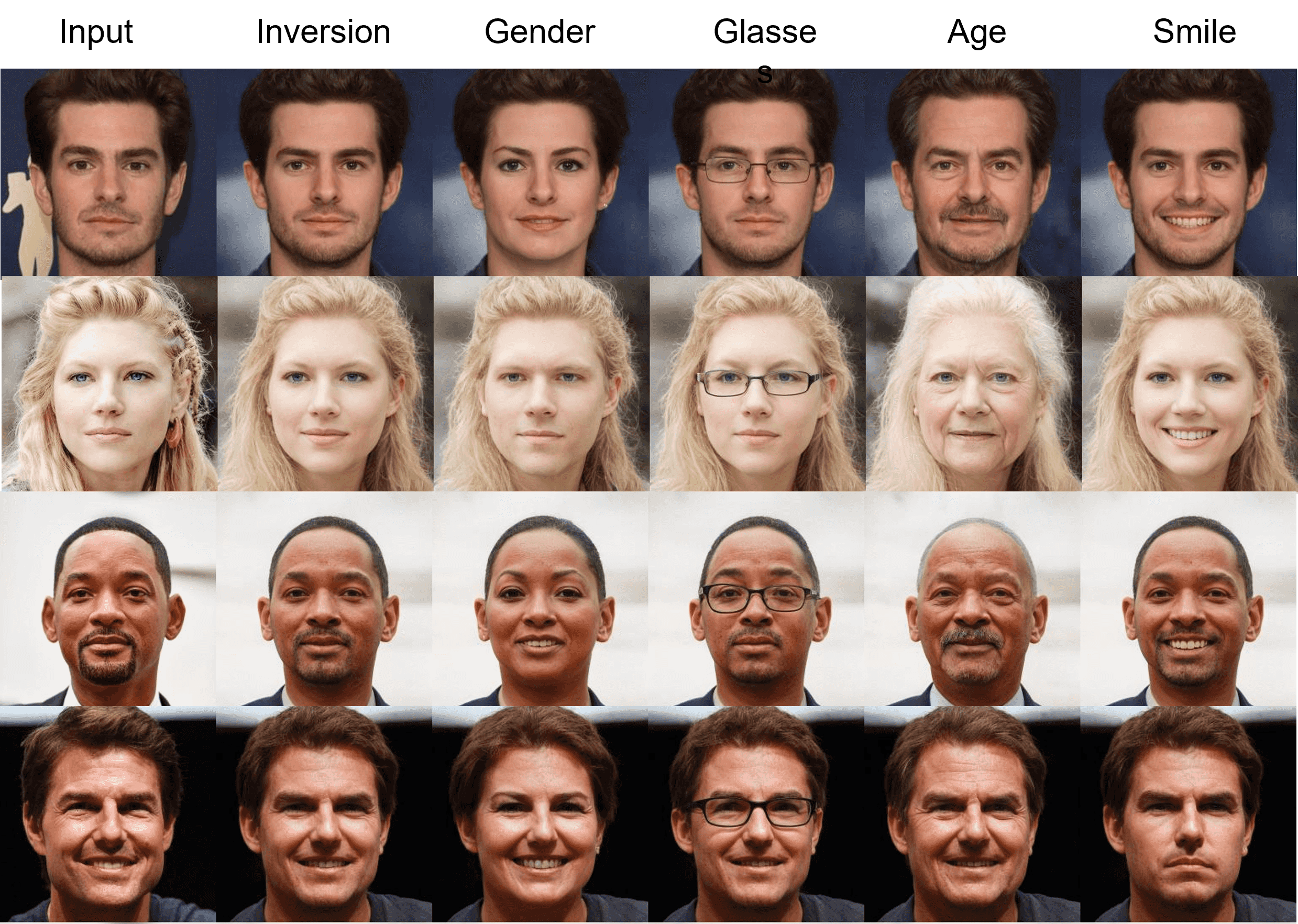}
   \caption{Manipulation of real images, gathered from CelebA-HQ \cite{karras2017progressive} and the web, using ID-Style.}
   \label{fig:real}
\end{figure}

Figure \ref{fig:real} demonstrates examples of editing facial attributes of real images. Firstly, source images have been selected from the CelebA-HQ dataset \cite{karras2017progressive} and the web. These images are then mapped into the $W^+$ latent space of StyleGAN2 using e4e \cite{tov2021designing}. Finally, these latent codes go through ID-Style to get the manipulated versions of the input image. The results reveal that the proposed method can achieve high-quality outputs with such inverted latent codes of real images. 

\begin{figure}
  \centering
   \includegraphics[width=\linewidth]{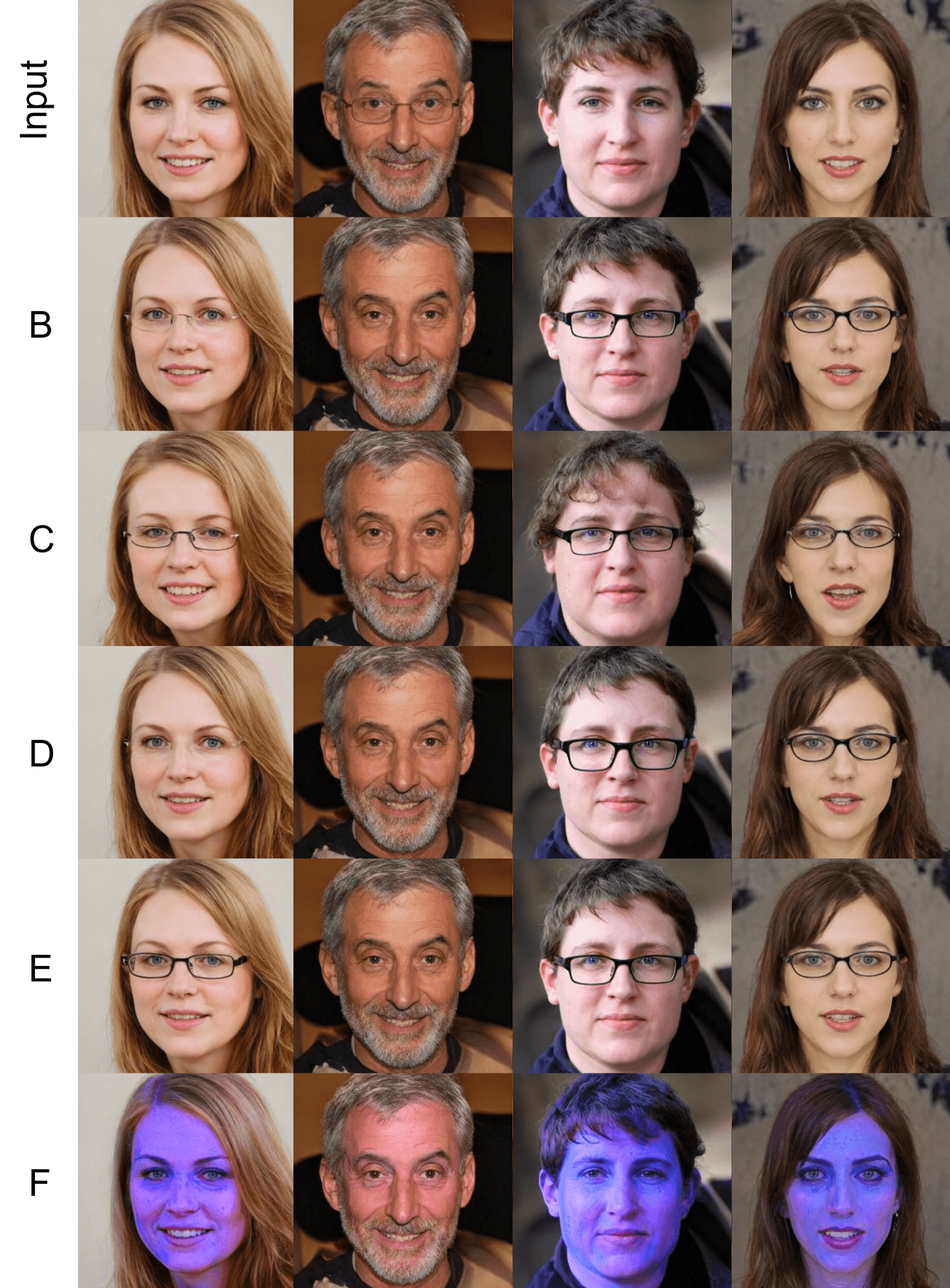}
   \caption{The impact of removing each ID-Style’s component on its performance in glasses manipulation. B-F are according to the settings of Table \ref{tab:ablation}.}
   \label{fig:ablation}
\end{figure}

\begin{figure*}[t]
  \centering
   \includegraphics[width=\linewidth]{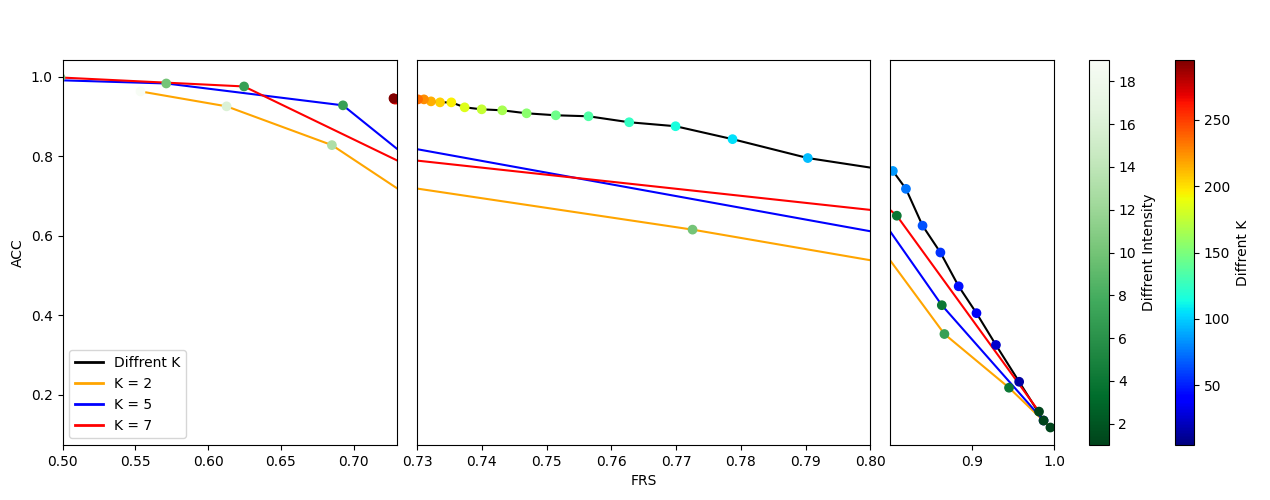}
   \caption{This plot depicts two experiments. The black line refers to the relation between top-k and the accuracy and FRS of manipulated images, in which k varies from 5 to 300 and is represented as colored dot points on the line. The other three lines refer to the experiment that investigates the possibility of decreasing top-k and consequently increasing the intensity of manipulation. Different colored lines are related to specific k, and the color of points on the lines shows the intensity.}
   \label{fig:topk}
\end{figure*}

\subsection{Sparsity-Entanglement Trade-off}
As mentioned in Section \ref{sec:method}, we claim that each attribute of the input image is affected by a separate subset of the elements of the global directions due to the sparsity loss of our model. To check if this statement is correct or not, the angle between every two global directions in $R^{512}$ is calculated. Figure \ref{fig:degre} shows the size of the angles between pairs of global directions in degrees. It is understood that whenever two vectors are orthonormal, the angle between them is 90 degrees. Clearly, all degrees reported in the Figure \ref{fig:degre} are close to 90 (between 80 and 97), which confirms our assumption. 

We even go beyond that and design another experiment to investigate the sparsity of global directions. In this experiment, we use only the top-k elements of $\Delta w$ to manipulate images, i.e., the $k$ elements of each row of $\Delta w$ with the largest absolute values. Figure \ref{fig:topk} shows the results of this experiment. It is clear that increasing $k$ beyond 150 does not have any considerable effect on results. In other words, only 150 elements of each global direction have effective values, and the rest are almost zero, which is equal to the sparsity of the global directions. This experiment also disclosed to us that there are many directions in StyleGAN2’s latent space for manipulating a specific attribute of an input image, and the best direction to preserve identity along manipulating the intended attribute is the one that engages around 150 elements of a 512-dimensional vector of the $W$ space with low intensities instead of those directions that affect few elements with higher intensities. 

To clarify the above assertion, the value of $k$ is decreased from 300 to 1, and the accuracy and FRS for different values of $k$ are shown in Figure \ref{fig:topk} (black line). As expected, decreasing $k$ increases FRS and reduces the accuracy of the model since fewer elements of the latent vector of the input image are altered, and the output is more similar to the input image. However, decreasing $k$ from 300 to 200 has a slight effect on both accuracy and FRS, while decreasing $k$ from 200 to 100 gradually deteriorates the accuracy and increases the FRS. For the $k$ values lower than 100, the accuracy of the model rapidly falls down. In the third experiment, $k$ is set to small values ($k=2,5,7$), and the intensity ranges from 1 to 20 in the hope of increasing the attribute change and improving the accuracy of the model. The results are shown in Figure \ref{fig:topk} for the three values of $k$ and different intensity values. As the results show, in lower ks', increasing the intensity improves the accuracy of manipulating attributes, but deteriorates the identity preservation. Generally, the best results are obtained for $100<k<200$. 

\subsection{Ablation Study}
\begin{table}
  \centering
  \small
  \caption{Ablation study on our proposed losses, embeddings, and MLPMixer-like component of IAIP. In each column, the best result is shown in bold and the worst result is underlined.}
  \begin{tabular}{l c c c c}
    \toprule
    Model & $\uparrow FRS$ & $\downarrow LPIPS$ & $\uparrow mACC$ \\
    \midrule
    A: ID-Style & 0.761 & 0.105 & \textbf{96.67} \\
    B: w/o $\mathcal{L}_{dir}$ & 0.756 & \textbf{0.102} & 95.82\\
    C: w/o $\mathcal{L}_{sparsity}$ & 0.778 & 0.104 & 95.98 \\
    D: w/o MLPMixer & 0.774 & 0.104 & 96.36 \\
    E: $w/o Embedding_{input}$ & \uuline{0.747} & 0.108 & 96.00\\
    F: $w/o Embedding_{output}$ & \textbf{0.801} & \uuline{0.121} & \uuline{88.51}\\
    \bottomrule
  \end{tabular}
  \label{tab:ablation}
\end{table}
In this subsection, we study the importance of each component of our method by removing or disabling one part of the model at a time. Table \ref{tab:ablation} and Figure \ref{fig:ablation} show the role of each part of ID-Style. Due to the complexity of learning to manipulate the Glasses attribute during training, in terms of identity preservation, we only illustrate the role of each component on this attribute in Figure \ref{fig:ablation}.  

Configuration B depicts the significance of the direction loss. Even though the LPIPS of model without this loss is slightly better than the full model, the precision of attribute manipulation is lower. Figure \ref{fig:ablation}-B also confirms this statement. Regarding the sparsity loss, it is obvious from Figure \ref{fig:ablation}-C that removing this loss function from the training of the model deteriorates the identity preservation and the quality of the output image, despite the evaluation metrics of Table \ref{tab:ablation} which show small changes. 

By disabling the MLPMixer part of the IAIP network, the identity preservation aspect of the model does not change noticeably, but the model completely fails to add glasses to some images (Figure \ref{fig:ablation}-D). 
Removing the positional embedding of the input layer causes the worst FRS value and deteriorates the LPIPS and mACC as well (Table \ref{tab:ablation}-E). The last rows of Table \ref{tab:ablation} and Figure \ref{fig:ablation} indicate the necessity of applying manipulation in specific layers of the latent vector. Removing layer embedding from the EDA component results in unwanted changes in the output image.

\section{Conclusion}
In this paper, we introduced a miniature neural network architecture to explore semantic directions in the latent space of StyleGAN2 under identity-preserving constraints. Results,  qualitatively and quantitatively, approved that the performance of ID-Style is far better than state-of-the-art methods in identity preservation and attribute controlling precision. Furthermore, we illustrated our hypothesis' correctness in designing the network through analytical experiments. To sum up, our experimental results show that ID-Style represents a competitive approach that utilizes a pre-trained and large-scale generator to manipulate image attributes. Future work should focus on adapting this small network in other generator models that have a latent space, such as diffusion models. 

{\small
\bibliographystyle{ieee_fullname}
\bibliography{egbib}
}

\end{document}